\documentclass[10pt,twocolumn,letterpaper]{article}

\usepackage{wacv}              % To produce the CAMERA-READY version
\usepackage{graphicx}
\usepackage{amsmath}
\usepackage{amssymb}
\usepackage{booktabs}
\usepackage{enumitem}

\usepackage[pagebackref,breaklinks,colorlinks]{hyperref}

\usepackage[acronym]{glossaries}

\newacronym{snn}{SNN}{Spiking Neural Network}
\newacronym{ann}{ANN}{Artifical Neural Network}
\newacronym{rnn}{RNN}{Recurrent Neural Network}
\newacronym{cnn}{CNN}{Convolutional Neural Network}
\newacronym{lif}{LIF}{Leaky Integrate-and-Fire}
\newacronym{plif}{PLIF}{Parametric Leaky Integrate-and-Fire}
\newacronym{stdp}{STDP}{Spike-timing-dependent plasticity}
\newacronym{mac}{MAC}{Multiply and Accumulate}
\newacronym{ac}{AC}{Accumulate}
\newacronym{flop}{FLOP}{FLoating Point Operation}
\newacronym{sgd}{SGD}{stochastic gradient descent}
\newacronym{gpu}{GPU}{Graphic Processing Unit}
\newacronym{mse}{MSE}{Mean Squared Error}
\newacronym{ssim}{SSIM}{Structural Similarity Index Measure}
\newacronym{map}{mAP}{mean Average Precision}
 
\usepackage{siunitx}
\newcommand\Tstrut{\rule{0pt}{2.4ex}}         % = `top' strut
\newcommand\Bstrut{\rule[-0.8ex]{0pt}{0pt}}   % = `bottom' strut
\usepackage{pifont}
\usepackage{multirow}
\usepackage{float}
\usepackage{xcolor}

\usepackage[toc,page]{appendix}

\usepackage[capitalize,nameinlink,sort&compress]{cleveref}
\crefname{section}{Sec.}{Secs.}
\Crefname{section}{Section}{Sections}
\Crefname{table}{Table}{Tables}
\crefname{table}{Tab.}{Tabs.}

\definecolor{somegray}{rgb}{0.5, 0.5, 0.5}
\newcommand{\darkgrayed}[1]{\textcolor{somegray}{#1}}
\makeatletter
\newcommand*\titleheader[1]{\gdef\@titleheader{#1}}
\AtBeginDocument{%
  \let\st@red@title\@title
  \def\@title{%
    \vskip-4em
    \bgroup\normalfont\large\centering\@titleheader\par\egroup
    \vskip1.5em\st@red@title}
}
\makeatother

\titleheader{\darkgrayed{This paper has been accepted for publication at the\\
IEEE Winter Conference on Applications of Computer Vision (WACV) 2025.}}

\def\wacvPaperID{2061} % *** Enter the WACV Paper ID here
\def\confName{WACV}
\def\confYear{2025}

\title{AnonyNoise: Anonymizing Event Data with Smart Noise\\to Outsmart Re-Identification and Preserve Privacy}
\author{\textsuperscript{1,2}Katharina Bendig
\quad
\textsuperscript{1,2}René Schuster
\quad
\textsuperscript{2}Nicole Thiemer
\quad
\textsuperscript{2}Karen Joisten
\quad
\textsuperscript{1,2}Didier Stricker
\\
\textsuperscript{1}DFKI -- German Research Center for Artificial Intelligence\\
{\tt\small \{katharina.bendig@, rene.schuster@, didier.stricker@\}dfki.de}\\
\textsuperscript{2}RPTU -- University of Kaiserslautern-Landau\\
{\tt\small \{n.thiemer@, karen.joisten@\}rptu.de}
}

\begin{document}

\maketitle

%%%%%%%%% ABSTRACT
\begin{abstract}
   The increasing capabilities of deep neural networks for re-identification, combined with the rise in public surveillance in recent years, pose a substantial threat to individual privacy. Event cameras were initially considered as a promising solution since their output is sparse and therefore difficult for humans to interpret. However, recent advances in deep learning proof that neural networks are able to reconstruct high-quality grayscale images and re-identify individuals using data from event cameras. In our paper, we contribute a crucial ethical discussion on data privacy and present the first event anonymization pipeline to prevent re-identification not only by humans but also by neural networks. Our method effectively introduces learnable data-dependent noise to cover personally identifiable information in raw event data, reducing attackers' re-identification capabilities by up to 60\%, while maintaining substantial information for the performing of downstream tasks. Moreover, our anonymization generalizes well on unseen data and is robust against image reconstruction and inversion attacks. \textbf{Code}: \url{https://github.com/dfki-av/AnonyNoise}
\end{abstract}

%%%%%%%%% BODY TEXT
\section{Introduction}
\label{sec:intro}

The amount of public and private surveillance increased strongly in the past decade as monitoring systems and cameras got cheaper and more energy efficient. The underlying motivation is often a heightening of security and the prevention of criminal acts. Additionally, there is a steady increase in the capabilities of deep learning methods for processing this visual data, allowing for the re-identification of individuals even between non-overlapping camera views \cite{ye2021deep}. This introduces a critical ethical debate about the value of individual privacy, which is endangered when high amounts of videos and images in public spaces are recorded, stored and processed. Moreover, in many cases the security of these cameras is lacking, potentially allowing unauthorized access and thus malicious usage of personal information. To enhance our understanding of data privacy from an ethical standpoint, we present a crucial discussion on the subject in \cref{sec:ethics}, providing valuable insights into the ethical dimensions of this complex issue.

\begin{figure}[t]
  \centering
  \includegraphics[width=0.79\linewidth]{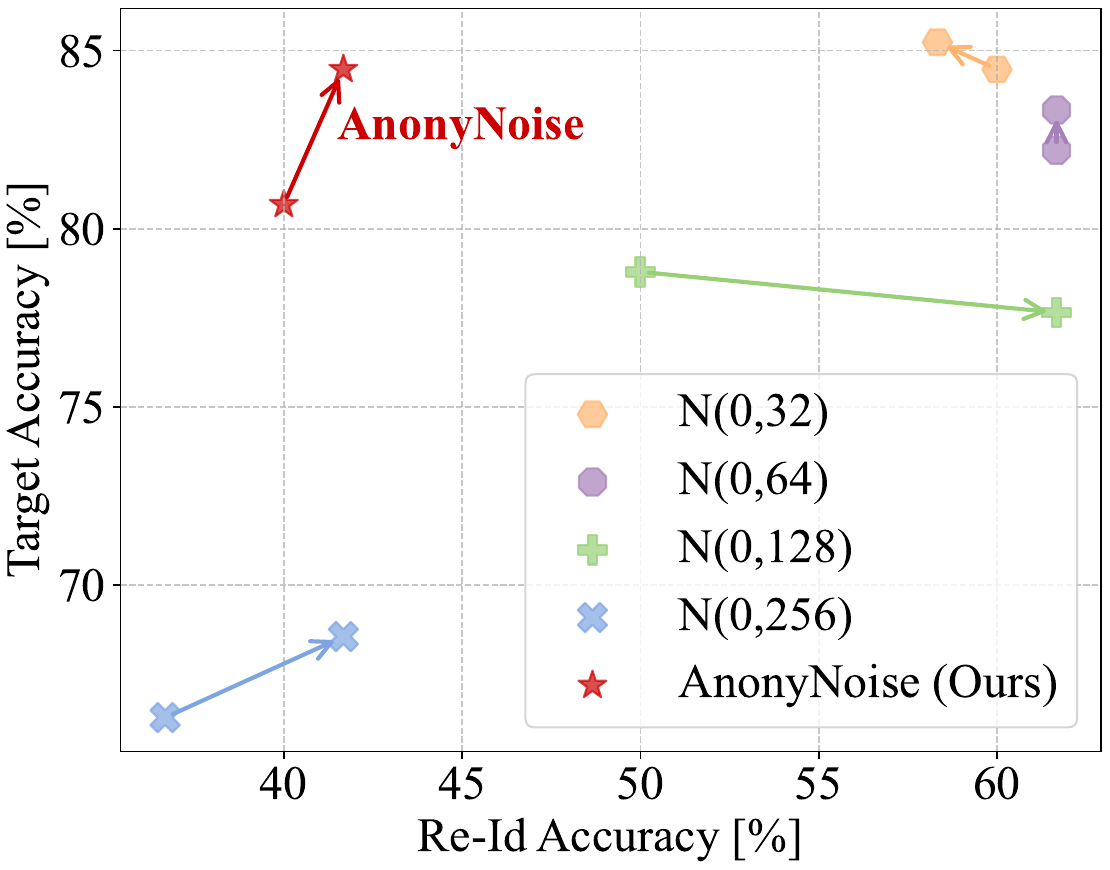}
  \caption{Visualization of re-identification and target accuracy for different noise applied to \textit{DVS-Gesture}\cite{dvsgesture}. The arrow is pointing from the result without to the result with a denoise network used. Our approach \textbf{AnonyNoise} is very robust against inversion.}
  \label{fig:teaser}
\end{figure}

Event cameras, gaining popularity as a low-energy alternative to RGB cameras, have recently been recognized for their effectiveness in always-on tasks like surveillance \cite{litzenberger2006estimation, rodriguez2020asynchronous, sultana2019iot}. A distinct feature of event cameras are their asynchronous pixels, which only register changes in intensity instead of absolute intensity values compared to RGB cameras. This allows for a high dynamic range as well as a high temporal resolution, making this technology attractive for many applications. Since event cameras register mainly movement information instead of visual details like in RGB frames, their output is difficult to comprehend for humans. This might seem to solve the privacy concerns for public surveillance, however latest research showed that neural networks are able to reconstruct grayscale images \cite{rebecq2019high, weng2021event, paredes2021back, zhu2022event} and perform re-identification \cite{ahmad2022event, Ahmad2024} based on event streams. This proofs the remainder of structural and personal information within event data, which can be exploited with the help of deep learning.

The work of \cite{ahmad2023person} went into a promising direction by training an anonymization network to prevent image reconstruction while enabling re-identification. However, it has to be argued, that re-identification is a main threat to the individual privacy. Their approach prevents human readability but does not prevent a re-identification by a neural network. This could potentially allow an attacker to track a person over multiple cameras despite the application of their anonymization approach. 

In our work, we expand the scope of event anonymization by introducing a novel training pipeline dedicated to preventing re-identification. Our approach goes beyond eliminating visual information essential for human interpretation or reconstruction and instead targets the removal of personal features exploitable by neural networks. This methodology offers a means to eliminate crucial personal information, particularly beneficial for future marketing applications where identification is unnecessary, such as gesture recognition and emotion detection in public spaces. Our method draws inspiration from the use of Gaussian noise, a straightforward technique for preventing re-identification. However, as demonstrated in \cref{fig:teaser}, the beneficial effects of noise can be undermined by the application of a denoising network. Additionally, while noise can cover identifying information, it also significantly degrades the accuracy of the target task. To overcome these challenges, we introduce \textbf{AnonyNoise}, a trainable data-dependent noise prediction technique. Applied to event data, \textbf{AnonyNoise} effectively reduces the risk of re-identification while preserving the critical information necessary for the desired target task.
We further enhance our method by employing adversarial training, ensuring that attackers cannot recover any remaining personally identifiable information. \textbf{AnonyNoise} thus achieves a sophisticated balance between privacy and performance, facilitating a more ethical and secure use of visual data in practical applications.

Our main contributions can be summarized in the following way:
\begin{itemize}[noitemsep,topsep=1pt,label={\tiny\raisebox{0.75ex}{\textbullet}},leftmargin=*]
    \item We provide a comprehensive ethical discussion of data privacy.
    \item We introduce the first trainable noise prediction network for re-identification prevention with event data. 
    \item Our adversarial training pipeline, ensures the robustness of our anonymization against exploitation from any re-identification network. 
\end{itemize}

\section{Ethics of Fragile Privacy}
\label{sec:ethics}

In the context of the ever-expanding technical possibilities of data collection, the protection of private data is of substantial importance. This imperative is underlined at the European legal level, encompassing the member states of the European Free Trade Association, with comprehensive guidelines in the General Data Protection Regulation (GDPR). Ethical implications regarding datasets and data quality are of growing importance which is emphasized in DEDA (Data Ethics Decision Aid) as well as in \cite{gebru2021datasheets}.

Central to this discourse is the ethical dimension of ensuring data protection at a fundamental technical level. It is crucial to acknowledge that data protection is not an isolated ethical value; rather, its significance lies in the potential misuse of data. The raison d'être for data protection is rooted in the meaning of data for the individual. The focus shifts from safeguarding data to safeguarding people, which is encapsulated in the words of Wiegerling \cite{Wiegerling2023}: "not data, but people need to be protected." This shift emphasizes the ethical concern surrounding the potential misuse of private data, framing the data protection issue as fundamentally tied to the importance of human privacy \cite{Wiegerling2023}.

Navigating the ethical horizon of surveillance capabilities unveils intricate challenges and the prospect of eroding privacy and private life \cite{nissenbaum2009privacy, bigdata-privacy-steinebach, behrendt2019privatsphare, rossnagel2022zukunft}. Surveillance options, while enhancing security in certain scenarios like criminal offenses, also pose a threat to the intrinsic human value of privacy. The dual-use character of technology becomes apparent, exemplified by the growing utilization of neuromorphic sensors in surveillance \cite{litzenberger2006estimation, rodriguez2020asynchronous, sultana2019iot}. 
Acknowledging this dual-use nature is crucial for addressing the ethical quandaries of surveillance, as it highlights the tension between enhanced security benefits and potential privacy risks.

The ethical discourse extends to products like event cameras, which as a sustainable technology (from the perspective of the data economy) warrant promotion. However, the ethical question of their dual-use character emerges, demanding careful consideration of potential implications. Critically addressing the ambivalent character of surveillance possibilities, the necessity for data anonymization methods arises, since this ambivalent character is allowing for “broader controls, but also deeper interventions” \cite{gaycken2013sicherheits}.

Privacy, from an ethical perspective, is the basis of data protection, particularly within the realm of \textit{informational privacy} \cite{seubert2023privatsphare}. Informational privacy refers the ability to control \textit{informational access} to sensitive data, underscoring the ethical challenges tied to preventing personal identification.

To preserve the human value of privacy, an interdisciplinary approach is warranted, intertwining ethical and technical perspectives at both foundational and application-specific levels.

\section{Related Work}
\label{sec:related_work}

\subsection{Privacy Preservation for RGB Data}
Person re-identification (ReId) describes the problem of retrieving a certain person either across non-overlapping camera views or for the same view but at different timestamps. As surveillance systems continue to expand, the volume of available data has increased significantly, deep learning has become a widely studied method for ReId applications using RGB images \cite{ye2021deep}. The success of these methods shows that machines are able to extract personal information rather easily from RGB images, which allows for an automatic and scalable exploitation of public surveillance data.
This poses a potential thread against the privacy of the individual as discussed in \cref{sec:ethics}.
To address these problems, multiple methods developed privacy-preserving solutions for RGB data. The authors of \cite{wu2020privacy} employ an anonymization network designed to modify RGB input data. This modified data still allows a target network to successfully perform action recognition, while simultaneously preventing identity detection by a privacy network. The pipeline is trained in an adversarial manner by alternating the optimization of the target/privacy-preserving and identification losses. SPAct \cite{dave2022spact} and TeD-SPAD \cite{fioresi2023ted} follow a similar pattern, but employ a contrastive self-supervised loss instead of focusing on preventing re-identification directly. The work of \cite{kumawat2022privacy} uses an encoder for the anonymization of RGB images, which produces outputs similar to event data. 
In contrast to these methods, we consider event data as input for the protection of privacy.

\subsection{Privacy Preservation for Event Data}
RGB data is naturally seen as a privacy concern, since their visual information can easily be understood by humans. Data from event cameras on the other hand are difficult to interpret for the human eye especially without any pre-processing. That is why event data was previously seen as privacy-preserving by default. However, recent deep learning methods like \cite{rebecq2019high, weng2021event, paredes2021back, zhu2022event} show, that high quality grayscale images can be recovered from event data.
Contrary to previous believes, these methods suggest that event streams contain personal information, that can be exploited by humans after being processed by a machine.
To enhance the privacy preservation, the authors of \cite{ahmad2023person} and \cite{Ahmad2024} suggest a pipeline for training an event anonymization network. This network is designed to anonymize the event input, preventing grayscale image reconstruction while maintaining the ability for re-identification. However, a notable limitation arises during their training process, where the parameters of the image reconstruction block remain untrained. Consequently, it might be possible to train a network with the capability to reconstruct images from the anonymized events. Moreover, their approach focuses only on the privacy preservation against the human perception while neural networks are still able to extract personal information for the re-identification. This poses a substantial threat to user privacy and presents an exploitable avenue for potential attackers. 
Our approach presents a comprehensive pipeline for event anonymization that prioritizes the removal of personal information exploitable by neural networks, while preserving usability for downstream tasks like action recognition.
Notably, our approach incorporates a min-max training strategy to ensure robustness against re-identification networks that undergo subsequent retraining on the anonymized event data.  

\begin{figure*}[htbp] 
    \centering

    \includegraphics[width=0.85\textwidth]{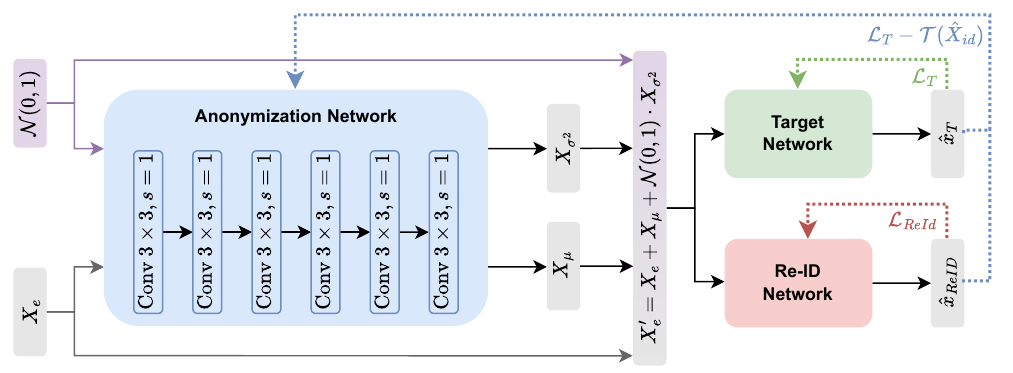}
       
    \caption{Our approach addresses all involved interests simultaneously. The target network (green) aims to optimize performance on the given target task, while a potential attacker (red) seeks to re-identify individuals from the input data. Our anonymization network (blue) tries to prevent this attack while still allowing for good performance of the target network. These two concurrent objectives the ethical trade-off between privacy and utility. The different losses of \cref{sec:method:pipeline} are employed to train each network according to its specific objective (indicated by dotted lines).}
    \label{fig:pipeline}
\end{figure*}

\section{Method}
\label{sec:method}
The goal of our anonymization method is to manipulate the event-based input in order to prevent re-identification while preserving the ability to perform downstream tasks based on the anonymized events. As shown in \cref{sec:reidattack}, Gaussian noise is a simple way to prevent re-identification based on event data. However, by applying a denoising network its effect can be partially diminished. Therefore, we are introducing a learnable Gaussian noise, which can not easily be removed and still prevents re-identification.
To do so, we propose a training pipeline as visualized in \cref{fig:pipeline}, which contains the following three main components: An \textit{anonymization network }$f_{Anon}$ to manipulate the raw event data, a \textit{re-identification network} $f_{ReId}$ as an adversarial trainer and a \textit{target network} $f_T$ to perform the downstream task, \eg action and emotion recognition.
These components are trained in three stages.
First, initial pre-training of individual models is performed. Second, the components are trained all together in an adversarial manner. Lastly, the weights of the anonymization network are frozen and the target and re-identification network are fine-tuned on the anonymized data, simulating a potential privacy attack.

\subsection{Event Data Representation}
The output of an event camera substantially differs from RGB cameras as their pixels register logarithmic intensity changes asynchronously instead of absolute intensity values at a frame rate. This results in a sparse stream of events in the form of tuples $e_i = (x_i, y_i, t_i, p_i)$, which include a pixel position $(x_i, y_i)$, a timestamp $t_i$ and a polarity $p_i \in \{0,1\}$, depicting a negative or positive change in intensity.

It is very challenging for neural networks to process such sparse input data, especially when modern clock-based \glspl*{gpu} are used during the training. For these reasons, we aggregate the event stream into dense event histograms $X_e$ of shape $(2 * T , H, W)$  with $(H,W)$ as the height and width of the event sensor and $2 *T$ channels for $T=5$ timesteps, which are split into the negative and positive polarity.

\subsection{Network Architectures and Pre-Training} \label{sec:method:architectures}
\paragraph{Anonymization Network.}
The anonymization of the event stream is done by the anonymization network $f_{Anon}$, which acts as as a noise generator in order to cover any privacy information inside the event data. 
The network's input are the event histograms $X_e$ concatenated with random noise $n = \mathcal{N}(0,1)$ in the same shape as $X_e$. The output is a per pixel mean $X_\mu$ and standard deviation $X_{\sigma^2}$, which are combined with the noise $n$ to form the anonymized data:
\begin{equation}
    X'_e = X_e + X_\mu + X_{\sigma^2} \cdot n.
\end{equation}
This is equivalent to the widely used re-parameterisation trick, which allows for the separation of the stochastic and deterministic processes in order to avoid the problem of back-propagating through a stochastic node. Moreover, we use the random noise as an input for $f_{Anon}$ to allow the network to learn the correlation between its output and the resulting anonymized data. 

In order to keep the architecture lightweight and applicable for mobile setting, we only utilize six convolutional layers with a $3\times 3$ kernel. This has the additional benefit that $f_{Anon}$ can be applied to any input resolution, allowing for generalisability to a variety of input data. Since event data is comparably sparse, the network needs the ability to create exact per pixel decisions for its output. For this reason, we utilize stride of 1 in all convolutional layers and thus avoid upsampling methods, which might result in additional blurring and loss of information. Notably, the network does not include skip connections in order to allow the network more flexibility in its noise prediction.

\paragraph{Re-Identification Network.}
Deep learning methods for re-identification aim to learn a distinctive feature representation for images of different individuals. This representation has to ensure that the features based on the same person are similar, while the features between different people exhibit dissimilarities.
Such a design allows for the matching of previously unseen validation samples in a query set to samples archived in a gallery. 
Inspired by \cite{ahmad2023person}, we choose a ResNet50 \cite{he2016deep} architecture for the backbone of our re-identification network $f_{ReId}$, and adjust the input weights for the $2*T$ channels of our event representation.

Moreover, we replace the last fully connected layer with a block comprising a linear layer, a batch normalization layer, a Leaky ReLU activation, and a dropout layer. This modification results in a 256-dimensional feature vector $\hat{X}_{id}$ for identification.

To train this model, we append another fully connected classifier layer with an output depending on the number of classes (different identities) in the training data.
Based on initial weights trained on ImageNet \cite{deng2009imagenet}, we further pre-train $f_{ReId}$ to perform re-identification on the raw event histograms of the used datasets. 

\paragraph{Target Network.}
The goal of our target network $f_T$ is to preserve information necessary for various downstream tasks during the anonymization training, The concrete task is therefore depended on the used dataset. In this work, we focus on gesture recognition and emotion detection, important tasks for modern human-machine-interaction that can be completely detached from sensitive personal information.
Regarding the network architecture, we utilize the same structure as for $f_{ReId}$ (with a different number of output channels for the classification layer, \ie the number of gestures or emotions) and apply a similar pre-training on the raw event histograms using a cross-entropy loss.

\subsection{Pipeline Training} \label{sec:method:pipeline}
Once the networks are pre-trained, we train all of them together as a pipeline in an adversarial end-to-end manner.
We strongly argue that simply fixing the weights of the re-identification network, $f_{ReId}$, during training is insufficient. In such a setup, the anonymization network, $f_{Anon}$, could merely learn to deceive a single instance of $f_{ReId}$, without ensuring robustness against other re-identification models that may still recover personal information from the anonymized data. To address this limitation, we adopt a min-max training strategy, where $f_{ReId}$ is continuously updated, allowing it to adapt its attacks in response to changes in the anonymization process. This dynamic interaction ensures that as $f_{Anon}$ evolves to anonymize the input data, $f_{ReId}$ concurrently adjusts to uncover any remaining identifiable information, resulting in a more robust anonymization system.

\paragraph{Re-Identification Loss.}
As discussed before, the objective of $f_{ReId}$ is to produce features, that have a high similarity for different images of the same person. Therefore a standard approach is the utilization of the identity loss, which includes the cross-entropy loss $\mathcal{CE}$ between the target and predicted identifier as well as the triplet loss $\mathcal{T}$ between all the features of the samples in one batch. The triplet loss minimizes the distances between features related to the same identity and maximizes the feature distance between different individuals. Therefore the loss to improve the re-identification attack is defined as:
\begin{equation}
    \mathcal{L}_{ReId} = \mathcal{CE}(x_{id}, \hat{x}_{id}) + \mathcal{T}(\hat{X}_{id}),
\end{equation}
with $x_{id},\hat{x}_{id}$ as the target and predicted identifier and $\hat{X}_{id}$ as the predicted features based on the anonymized event input $X'_e = f_{Anon}(X_e)$. 

\paragraph{Target Loss.}
For the chosen target downstream task, we apply the cross-entropy between the ground truth $x_T$ and predicted label $\hat{x}_T$ of the target task, \ie the gesture or emotion category, that is based on the anonymized input $X'_e$:

\begin{equation}
    \mathcal{L}_{T} = \mathcal{CE}(x_T, \hat{x}_T).
\end{equation}

\paragraph{Anonymization Loss.}
The target network $f_{T}$ shares the objective of a correct target prediction with the anonymization network $f_{Anon}$. Thus, we incorporate the loss term $\mathcal{L}_T$ into the training of $f_{Anon}$. Additionally, the anonymized output from $f_{Anon}$ should lead to an incorrect re-identification when processed by $f_{ReId}$. To enforce this, we apply a negative triplet loss, ensuring that features of the same individual diverge, thereby lowering the re-identification capabilities.

In summary, we utilize the following loss for the training of $f_{Anon}$:
\begin{equation}
    \mathcal{L}_{Anon} = \mathcal{L}_T - \mathcal{T}(\hat{X}_{id}).
\end{equation}

\subsection{Post-Training Attack} \label{sec:method:attack}
After adversarial training for privacy preservation, we simulate a hypothetical iterative re-identifaction attack by training $f_{ReId}$ again on the anonymized and fixed event representations $X'_e$, \ie $f_{Anon}$ is frozen and not updated anymore.
Similarly, we fine-tune the target network on the anonymized data in this post-training stage.

\section{Experiments and Results}
\label{sec:experiments_and_results}

We conduct a series of experiments to demonstrate the effectiveness of our anonymization strategy and provide insights into its robustness and generalization capabilities.
After introducing the used datasets, we specify implementation and training details to ensure reproducibility. 
Next, results after pre-training and after the re-identification attack (post-training) are presented and compared.
On data that only includes re-identification labels and no additional information for any downstream task, we validate the generalization of the anonymization network.
Lastly, experiments and results regarding the robustness are discussed. 

\subsection{Datasets}
In order to validate our approach, we apply our anonymization pipeline on the real event datasets \textit{DVS-Gesture} \cite{dvsgesture} and \textit{SEE} \cite{see} and test the anonymization capabilities against re-identification. Additionally, we show the generalization of the anonymization on the \textit{Event-ReId} \cite{ahmad2023person} dataset. 

\textit{DVS-Gesture} \cite{dvsgesture} is a gesture recognition dataset, that was directly recorded with an event camera. The dataset includes 29 different subjects, 6 in the test and 23 in the training split. The subjects perform 11 hand gestures, which results overall in 1342 event samples with a size of $128px \times 128px$. We constructed a query set, by choosing a random sample per subject and per target label. During the re-identification validation, we filter out samples from the gallery dataset with both the same subject and target label as the query sample. 

\textit{SEE} \cite{see} is a single-eye emotion recognition dataset captured using a head-mounted event camera. It records seven distinct emotions: anger, happiness, sadness, neutrality, surprise, fear, and disgust. Each sample is captured at a resolution of $180px \times 180px$ and consists of multiple event frames. We construct event histograms by partitioning all the event frames temporally into $T=5$ timesteps and aggregating them for each polarity. To tailor the dataset for the re-identification task, we first exclude subjects who did not have recordings for all seven emotions. Next, we select subjects with the fewest samples to form the validation set. This process resulted in a split of 78 subjects for training and 20 subjects for validation, with each validation subject having only one sample per emotion. For the re-identification task, two random samples are selected per subject for the query set, with the remaining samples assigned to the gallery set. 

\textit{Event-ReId} \cite{ahmad2023person} is an event dataset for the re-identification of 33 subjects, 11 in the test split and 22 in the training split. The subjects are recorded walking by 4 event cameras, which have a non-overlapping field of view. Following the processing in \cite{ahmad2023person}, we construct event samples for time windows of 33.3ms and further resize them to a $304px \times 304px$ resolution. Since the used query/gallery split is not provided, we construct our own split using the same random process as described in \cite{ahmad2023person}. 

\subsection{Implementation}
In order to ensure optimal training settings, we utilize different training parameters for the three datasets, which are listed in more detail in the supplementary material.
For the pipeline training, we choose a higher learning rate and a step-wise decay each 100 epochs for all the auxiliary networks as it is often done in adversarial training settings. This helps provide a strong training signal to the anonymization network throughout the whole training process. As an optimizer we use AdamW \cite{AdamW}.

During training on both datasets, we apply geometrical augmentation as it is a widely established technique. Following the examples of \cite{eventmix, ShapeAug}, we use random cropping after zero-padding by a twelfth of the original size, random horizontal flipping and random rotation by up to $15^{\circ}$. Notably, we do not apply horizontal flipping for \textit{DVS-Gesture} because the labels are dependent on the direction of the gestures. 

\begin{figure*}[t]
  \centering

  \begin{subfigure}[c]{0.175\textwidth}
    \includegraphics[width=\linewidth]{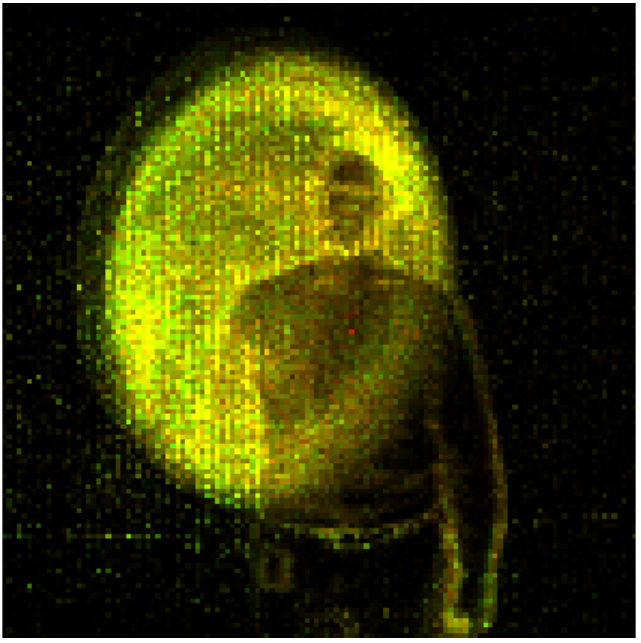}
    \caption{\textit{DVSG} Original}
    \label{fig:dvsg:orig}
  \end{subfigure}
  \hfill
  \begin{subfigure}[c]{0.175\textwidth}
    \includegraphics[width=\linewidth]{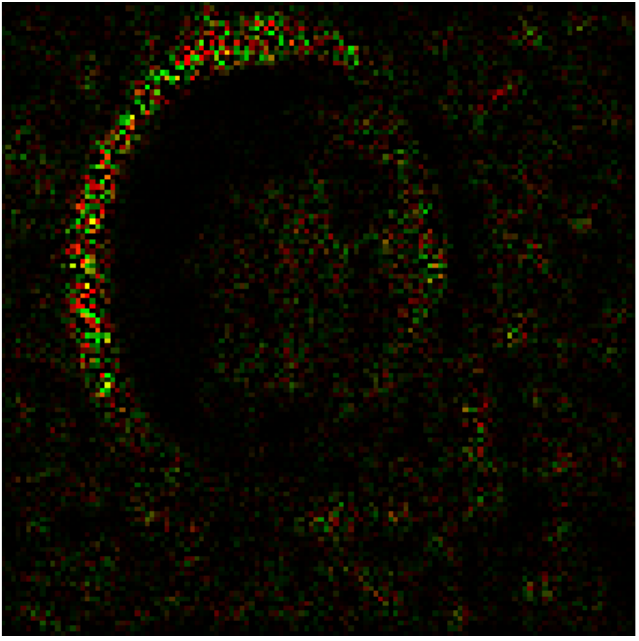}
    \caption{\textit{DVSG} Anonymized}
    \label{fig:dvsg:anon}
  \end{subfigure}
  \hfill
  \begin{subfigure}[c]{0.175\textwidth}
    \includegraphics[width=\linewidth]{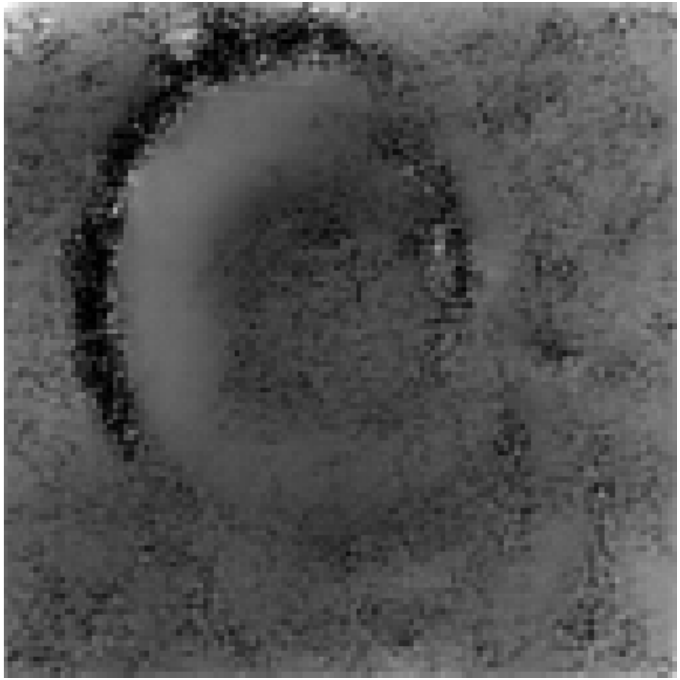}
    \caption{\textit{DVSG} Image recon}
    \label{fig:dvsg:recon}
  \end{subfigure}
  \hfill
  \begin{subfigure}[c]{0.175\textwidth}
    \includegraphics[width=\linewidth]{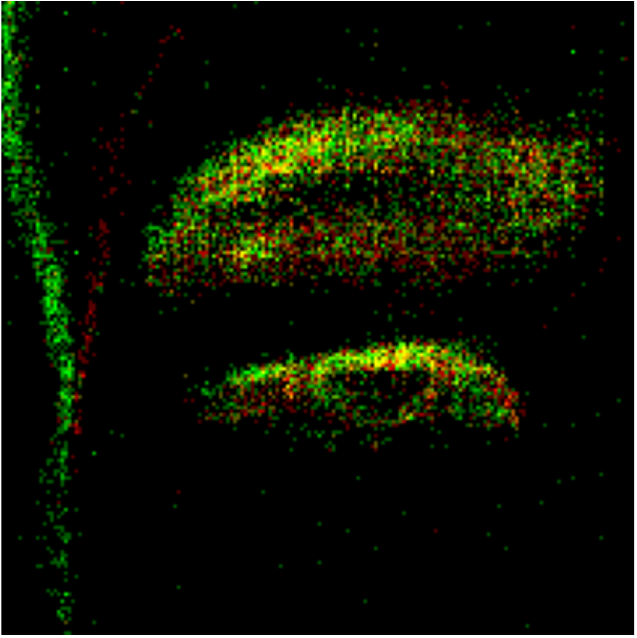}
    \caption{\textit{SEE} Original}
    \label{fig:see:orig}
  \end{subfigure}
  \hfill
  \begin{subfigure}[c]{0.175\textwidth}
    \includegraphics[width=\linewidth]{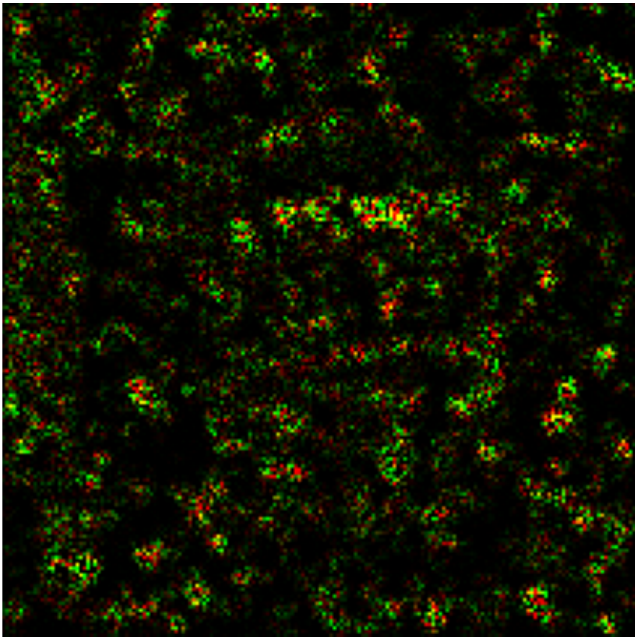}
    \caption{\textit{SEE} Anonymized}
    \label{fig:see:anon}
  \end{subfigure}

  \caption{Example visualizations of the raw and anonymized events for \textit{DVS-Gesture} \cite{dvsgesture} and \textit{SEE} \cite{see}. Subfigure c) depicts the result of grayscale image reconstruction from the anonymized data using E2VID \cite{rebecq2019high}. The images are visually enhanced for human perception.}
  \label{fig:vis_events}
\end{figure*}

\subsection{Re-Identification Attack and Target Task} \label{sec:reidattack}

\begin{table*}[t]
\centering
\caption{Evaluation of the target task and the re-identification on DVS-Gesture \cite{dvsgesture} and SEE \cite{see} for different training stages and the application of standard Gaussian noise.}
\label{tab:pretrain}
\begin{tabular}{ccc|c|c|c} 
  Dataset & Input & Stage  & $acc_T [\%] \uparrow$  & $acc_{id} [\%] \downarrow $ & $mAP_{id} [\%] \downarrow$\Bstrut\\
\hline 
\multirow{6}{*}{DVS-Gesture \cite{dvsgesture}} & Raw & Pre-training  & 92.42 & 98.33 & 69.64\Tstrut\\
 & Noise std=32 & Post-training  & 84.47 & 60.00 & 27.24\Tstrut\\
 & Noise std=64 & Post-training  & 82.20 & 61.67 & 27.04\Tstrut\\
 & Noise std=128 & Post-training  & 78.79 & 50.00 & 23.40\Tstrut\\
 & Noise std=256 & Post-training  & 66.29 & 36.67 & 20.07\Tstrut\\
 & Anonymized & All stages & 80.68 & 40.00 & 20.69\Bstrut\\
\hline
\multirow{5}{*}{SEE \cite{see}} & Raw & Pre-training  & 77.14 &  72.5 & 43.27\Tstrut\\
 & Noise std = 1 & Post-training  & 66.43 & 47.5 & 25.63\Tstrut\\
 & Noise std = 2 & Post-training  & 60.00 & 35.00 & 17.98\Tstrut\\
 & Noise std = 4 & Post-training  & 55.00 & 32.5 & 17.08\Tstrut\\
 & Anonymized & All stages & 55.71 & 25.0& 13.14\Bstrut\\
\end{tabular}
\end{table*}

During the pre-training, we train the auxiliary networks $f_{T}$ and $f_{ReId}$ for their respective tasks on the raw event data in order to provide a better initialization and therefore a more meaningful training signal during the pipeline and post-training. 

The results of the pre-trained target and re-identification networks on \textit{DVS-Gesture} \cite{dvsgesture} and \textit{SEE} \cite{see} are given in \cref{tab:pretrain}. Both networks are evaluated in terms of accuracy (${acc}_T$ and ${acc}_{id}$) and for $f_{ReId}$ we also compute the \gls*{map} of identity retrieval from the validation dataset. Both networks achieve a reasonable result on the raw event data.

We further demonstrate the effect of Gaussian noise with varying standard deviations for both datasets in \cref{tab:pretrain}. As the results indicate, introducing noise significantly enhances anonymization by substantially reducing re-identification accuracy and mAP. This effect arises from the distribution of events, where regions of interest with high motion typically contain a dense concentration of events. Consequently, noise effectively covers personally identifiable features while still allowing for target detection. However, the optimal standard deviation depends heavily on the specific dataset, as excessive noise inevitably also decreases the target task performance, limiting its practical applicability.

Utilizing the pre-trained weights of the auxiliary networks, we then continue with our strategy of adversarial pipeline training as described in \cref{sec:method:pipeline}. The results in \cref{tab:pretrain} show, that our method effectively prevents re-identification reducing ${acc}_{id}$ by nearly 60\% for \textit{DVS-Gesture} and about 50\% for \textit{SEE}, which is considerably lower than most applications of noise. Moreover, our approach is able to almost maintain the target accuracy ${acc}_T$ only reducing it by 12\% for \textit{DVS-Gesture} and 22\% for \textit{SEE}, vastly outperforming all noise with a similar re-identification prevention. Example visualizations for both datasets can be found in \cref{fig:vis_events}

Overall, the results proof, that our pipeline training is capable to anonymize events effectively and is indeed able to cover minute private information, while preserving necessary clues for other downstream tasks.

\subsection{Generalization of Protection}
\label{sec:evreid}

\begin{table*}
\centering
\caption{Re-identification results on Event-ReId \cite{ahmad2023person} for raw and anonymized event input based on the anonymization network trained for \textit{DVS-Gesture} \cite{dvsgesture} without further fine-tuning.}
\label{tab:reid}
\begin{tabular}{c|c|c|c|c}
 Method & top-1 acc [\%] $\downarrow$ & top-5 acc [\%] $\downarrow$ & top-10 acc [\%] $\downarrow$ & mAP [\%] $\downarrow$\Bstrut\\
\hline 
No Privacy & 59.09 & 77.27 & 81.82 & 34.04\Tstrut\\
AnonyNoise (Ours) & 38.64 & 61.36 & 68.18 & 15.23\\
\end{tabular}
\end{table*}

We evaluate the generalizability of $f_{Anon}$ on \textit{Event-ReId} \cite{ahmad2023person} without fine-tuning on this dataset. Since the dataset does not contain data for a downstream task, we utilize the anonymization network trained on \textit{DVS-Gesture} \cite{dvsgesture} and keep them fixed during the post-training for re-identification on \textit{Event-ReId}. 

The result of our experiments can be found in \cref{tab:reid}. They show that our anonymization method is able to decrease the re-identification accuracy of the retrained $f_{id}$ network by 20\% and more than half the $mAP$ score, all without fine-tuning $f_{Anon}$. This is especially remarkable since the two datasets are very different in terms of event density as well as in the human poses. While \textit{DVS-Gesture} contains only front-facing subjects, \textit{Event-ReId} includes recordings from diverse camera angles, further underscoring the robustness and generalization of our anonymization approach. 

\subsection{Ablation Study}

\begin{table*}
\centering
\caption{Results of the target task and the re-identification after post-training when we ablate and vary the input and output of $f_{Anon}$ during our pipelined training.}
\label{tab:ablation}
\begin{tabular}{cc|c|c|c}
 Input & Output   & $acc_T [\%] \uparrow$ & $acc_{id} [\%] \downarrow$ & $mAP_{id} [\%] \downarrow$\Bstrut\\
\hline 
Pre-Training & - & 92.42 & 98.33 & 69.64\\
$X_e$ & $X_\mu$ & 56.44 & 35.0  & 19.48\\
$X_e$ & $X_e + X_\mu$ &  81.82 & 83.33 & 45.48 \\
$X_e$ &$X_e +  n \cdot X_{\sigma^2}$   & 84.85 & 48.33 & 24.71 \\
$X_e, n$ &$X_e +  n \cdot X_{\sigma^2}$   & 80.68 & 40.00  & 20.69 \\
$X_e, n$& $X_e +  X_\mu + n \cdot X_{\sigma^2}$  & 82.58  & 45.00 & 22.75 
\end{tabular}
\end{table*}

We conduct an ablation study on different input-output configurations for the anonymization network $f_{Anon}$, as shown in \cref{tab:ablation}. For the input, we evaluate the use of raw events alone, as well as the concatenation of raw events with Gaussian noise $n = \mathcal{N}(0,1)$, which is applied to the network's output. For the output, we explore the addition of the noise $n$ or the raw events $X_e$ to the network output as well as scaling $n$ by predicting a pixel-wise standard deviation $X_{\sigma^2}$.

The results indicate that without the addition of the original input $X_e$ to the network's output, the re-identification accuracy indeed reduces drastically. However, the target accuracy is substantially decreases as well, as the network output lacks constraints tied to the input events. On the other hand, merely adding the network's output to the input event is resulting in limited anonymization, reducing $acc_{id}$ by only 15\%.

Introducing noise $n$ to the output significantly enhances anonymization, further lowering $acc_{id}$ by 35\%. This demonstrates the effectiveness of randomness, which poses a challenge for neural networks during re-identification. The best results are achieved when the used noise is additionally fed into the anonymization network as an input, achieving an overall reduction of the re-identification accuracy of almost 60\%. The noise input enables the network to determine its output in a data-dependent manner, including the computation of $X_{\sigma^2}$ based on both the noise and input event histograms. 

However, predicting and adding an additional $X_\mu$ does not yield further improvements, showing that maintaining a mean of zero is crucial for the effective application of this noise.

\begin{figure}[ht]
    \centering

    \includegraphics[width=\linewidth]{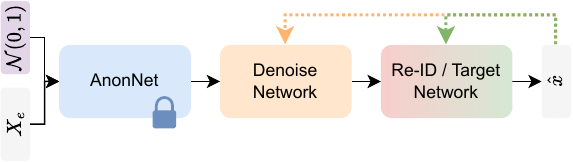}
       
    \caption{For the inversion attack, we insert a denoising network (orange) between the frozen anonymization network (blue) and the classification network (red/green) for the post-training.}
    \label{fig:posttraindenoise}
\end{figure}

\subsection{Robustness against Image Reconstruction and Inversion Attacks} \label{sec:experiments_and_results:inversion}

We evaluate our anonymization method against two more privacy attacks: Inversion and image reconstruction. 

An inversion attack utilizes a neural network to reverse anonymization by reconstructing the original data from its anonymized counterpart. To achieve this, we introduce a denoising network $f_{Denoise}$ and position it before the re-identification or target network, as illustrated in \cref{fig:posttraindenoise}, during post-training. While freezing the weights of the anonymization network, the denoising and target networks are trained jointly using a combination of the classification loss and the MSE loss between the original input event data $X_e$ and the reconstructed output $f_{Denoise}(X'_e)$. For the denoise network $f_{Denoise}$ we choose a simple architecture consisting of 15 convolutional layers ($k = 3 \times 3, s = 1$) with batch normalization. The results of this attack applied to our method as well as to standard Gaussian Noise are visualized in \cref{fig:teaser} and are listed in the supplementary material. The data shows, that the inversion is especially effective for noise with a high standard deviation, where it increases the re-identification accuracy by up to 11 \%. Our method, on the other hand, proofs to be very robust and even after the inversion attack retains a similar $acc_{id}$, while increasing the desired target accuracy by almost 4 \%.

Image reconstruction based on the events could be another way to regain private information after the anonymization. To evaluate this, we utilize a pre-trained E2VID network \cite{rebecq2019high}, a recurrent neural network for high-quality grayscale images reconstruction from event data. The qualitative result of the image reconstruction attack in \cref{fig:dvsg:recon} shows that the attack failed for DVS-Gesture \cite{dvsgesture}. Therefore, our anonymization is robust against possible grayscale reconstructions. Further reconstruction examples are provided in the supplementary material.

\section{Limitations}
While our event anonymization pipeline has proven effective on the utilized datasets, our current framework exclusively supports event histograms as the input representation. This is due to the computational infeasibility of processing sparse event streams with state-of-the-art neural networks on recent \glspl*{gpu}, which have a clock-based architecture. As improvements in sparse data processing continue, our approach stays flexible and provides potential avenues for future research extensions.

\section{Conclusion}
In this paper, we present an ethical analysis of data privacy and propose a novel anonymization method for event data for the prevention of re-identification risks. We argue that mere anonymization against human recognition based on visual features falls short in many applications. Instead, a more comprehensive approach is needed - one that anonymizes all personal information embedded in event data, preventing misuse by deep learning models.
Our proposed method achieves anonymization by predicting and applying noise in a data-dependent manner. When applied to event data, our approach effectively prevents re-identification attempts by neural networks while preserving the performance of downstream tasks. To demonstrate its effectiveness, we simulate a re-identification attack by training a neural network to identify individuals from the anonymized event data. The results of our experiments confirm that our method is robust against such attacks, while maintaining strong performance on target tasks such as gesture recognition.
Furthermore, our approach generalizes well to unseen data and remains robust against inversion and image reconstruction attacks. 
By covering unnecessary personal data, our method supports the ethical use of vision sensor data, ensuring secure storage and safeguarding privacy even amidst security threats, making it applicable to any recognition system where person identification is undesirable.

\section*{Acknowledgements}
This work was partially funded by the Carl Zeiss Stiftung Germany under the Sustainable Embedded AI project (P2021-02-009), and partially under the EU project dAIEDGE (GA Nr 101120726).
%%%%%%%%% REFERENCES
{\small
\bibliographystyle{ieee_fullname}
\bibliography{egbib}
}

%\clearpage
\appendix
\section*{Supplementary Material}
\section{Overview}

In this supplementary material, we provide a more detailed overview of AnonyNoise, a method developed for predicting data-dependent noise aimed at preventing re-identification. The document is structured as follows:
First, we detail the training parameters used in our implementation for the three datasets: DVS-Gesture \cite{dvsgesture}, SEE \cite{see}, and Event-ReId \cite{ahmad2023person}, in order to ensure reproducibility.
Next, we present numerical results from an inversion attack on our method, comparing its effectiveness to Gaussian noise when evaluated using a denoising network. This comparison provides insight into the robustness of AnonyNoise in contrast to traditional noise techniques in preventing data recovery and re-identification attempts.
We moreover provide our statement regarding our responsibility to human subjects in the datasets used during our experiments.
Lastly, we include an expanded set of visual examples across all datasets, including the results from image reconstruction attacks.

\section{Implementation Details}
The training parameters for each dataset are listed in \cref{tab:parameter}. Notably, we choose a higher learning rate for all the auxiliary networks during the pipeline training. This ensures that a strong training signal is consistently delivered to the anonymization network throughout the entire process. Moreover, we apply a step-wise learning rate decay every 100 epochs, with a gamma value of 0.5. For all networks and training stages, we employ the AdamW optimizer \cite{AdamW}. Additionally, the same parameter settings are used in the post-training phase for the inversion attack when utilizing the denoise network. Since the SEE \cite{see} dataset results in event histograms with a lower amount of events, we utilize a $\mathcal{N}(0,0.1)$ distribution for the added noise instead of $\mathcal{N}(0,1)$ for DVS-Gesture \cite{dvsgesture}.

\section{Inversion Attacks}

Reconstructing the original data from anonymized events is a key threat that any anonymization method must prevent. To evaluate this, we simulate an inversion attack by inserting a denoise network between the anonymization and classification networks. We compare the impact of this attack on the AnonyNoise method and Gaussian noise (with varying standard deviations) using the DVS-Gesture dataset \cite{dvsgesture}. The results are shown in \cref{tab:posttraindenoise} and correspond to Fig\onedot 1 of the main paper.

The data indicates that increasing the standard deviation of Gaussian Noise reduces re-identification accuracy. However, this also makes inversion easier, which increases the identity accuracy ($acc_{id}$) by more than 11\%. Furthermore, using noise with a high standard deviation significantly decreases the target accuracy, making this anonymization method unsuitable for practical applications.

In contrast, when AnonyNoise is applied, the denoise network is unable to increase the re-identification accuracy significantly. Instead, the inversion process positively impacts target accuracy, improving it by almost 4\%, a desirable outcome. This demonstrates that our method is robust against inversion attacks, even without explicitly optimizing for this feature during training.

\begin{table}[H]
\centering
\caption{Results of the inversion attack during post-training based on \textit{DVS-Gesture} \cite{dvsgesture}. In brackets is the difference to the results of the post-training without the inversion attack.}
\label{tab:posttraindenoise}
\begin{tabular}{cccc|c|c}
 Method & $acc_T [\%] \uparrow$ & $acc_{id} [\%] \downarrow$\Bstrut\\
\hline 
Raw       & 92.42 & 98.33\Tstrut\\
% $\mathcal{N}(0,32)$  & 84.47 (+0.00)& 58.33 (-1.67)\\
% $\mathcal{N}(0,64)$  & 82.95 (+0.75) & 60.00 (-1.67)\\
% $\mathcal{N}(0,128)$ & 76.89 (-1.90) & 56.67 (+6.67)\\
% $\mathcal{N}(0,256)$ & 67.42 (+1.13)& 43.33 (+6.66)\\
% AnonyNoise (Ours)    & 84.85 (+4.17)& 40.00 (+0.00)\\
$\mathcal{N}(0,32)$  & 85.23 (+0.75) & 58.33 (-1.67) \\
$\mathcal{N}(0,64)$  & 83.33 (+1.13) & 61.67 (+0.00)\\
$\mathcal{N}(0,128)$ & 77.65 (-1.14) & 61.67 (+11.67)\\
$\mathcal{N}(0,256)$ & 68.56 (+2.27) & 41.67 (+5.00) \\
AnonyNoise (Ours)    & 84.47 (+3.79) & 41.67 (+1.67)\\
\end{tabular}
\end{table}

\begin{table*}[t]
\centering
\caption{Training parameters for pre- and post-training as well as for training the anonymization pipeline for each dataset.}
\label{tab:parameter}

\begin{tabular}{cc|c|c|c|c c}

Dataset              & Training Phase            & Network       & Batchsize & Learning Rate & Scheduler & Epochs\Bstrut\\
\hline
\multirow{3}{*}{\begin{tabular}{c}DVS-Gesture \cite{dvsgesture}\\SEE \cite{see}\end{tabular}} & Pre/Post                  & Re-ID /Target Net & 32        &  $1\times 10^{-4}$   &  Cosine decay        & 200\Tstrut\\
                     & \multirow{2}{*}{Pipeline} & Re-ID /Target Net & 32        &  $1\times 10^{-3}$             &  Step-wise & 300\Bstrut\\
                     &                           & Anon Net      & 32        &  $5\times 10^{-4}$             &  Cosine decay         & 300\Bstrut\\
\hline
Event-ReId \cite{ahmad2023person}                   & Pre/Post                 & Re-ID Net        & 24        &  $1\times 10^{-3}$            &     Cosine decay       &    200\Tstrut\\
\end{tabular}
\end{table*}

\section{Responsibility to Human Subjects}
The datasets used in our experiments were obtained through collaboration with human subjects on a voluntary basis. The authors of the respective papers are accountable for ensuring that all essential consents were secured before the publication of the datasets. It is crucial to note that the data encompasses personally identifiable information, a regrettable but imperative aspect for our research aimed at training networks for re-identification prevention.

\section{Extended Visualizations and Image Reconstruction}

\begin{figure*}[t]
  \centering
  \begin{subfigure}[c]{0.03\textwidth}
    \rotatebox[origin=c]{90}{DVS-Gesture}
  \end{subfigure}%
  \begin{subfigure}[c]{0.185\textwidth}
    \includegraphics[width=\linewidth]{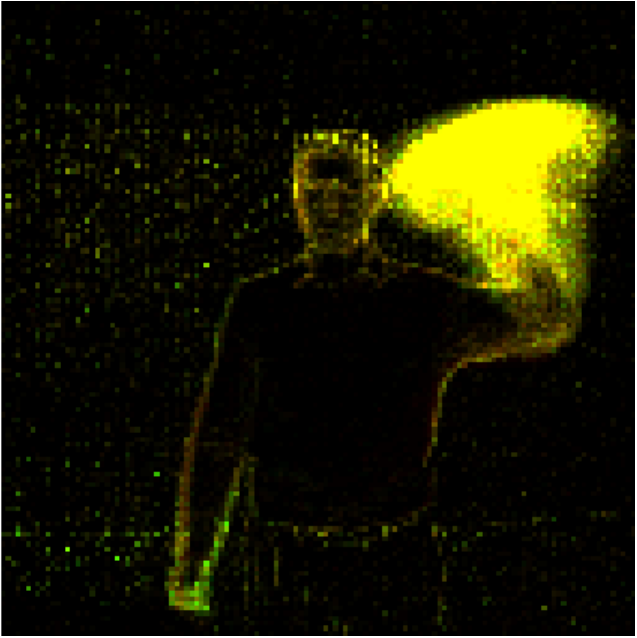}
  \end{subfigure}
  \begin{subfigure}[c]{0.185\textwidth}
    \includegraphics[width=\linewidth]{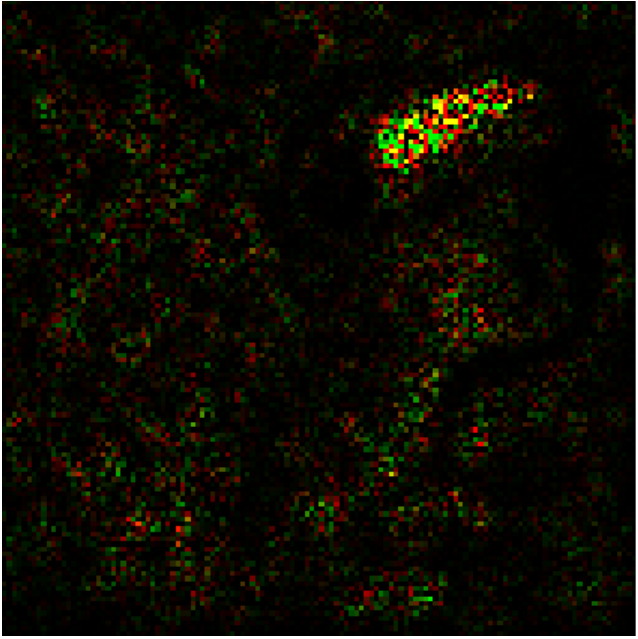}
  \end{subfigure}
  \begin{subfigure}[c]{0.185\textwidth}
    \includegraphics[width=\linewidth]{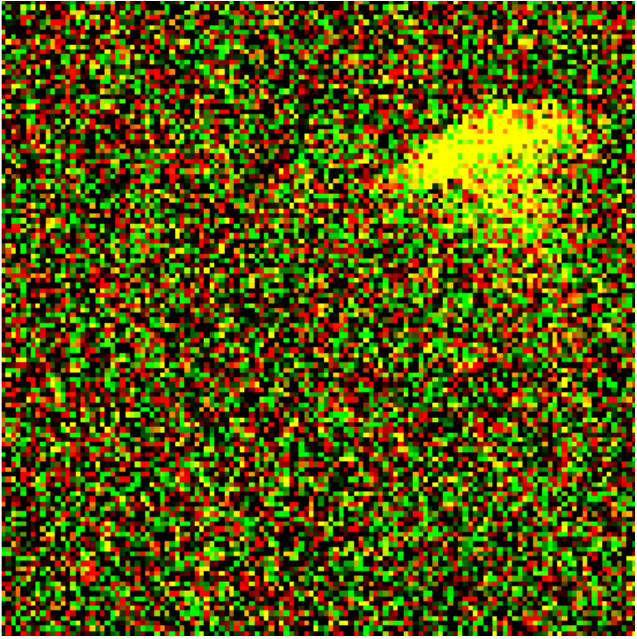}
  \end{subfigure}
  \begin{subfigure}[c]{0.185\textwidth}
    \includegraphics[width=\linewidth]{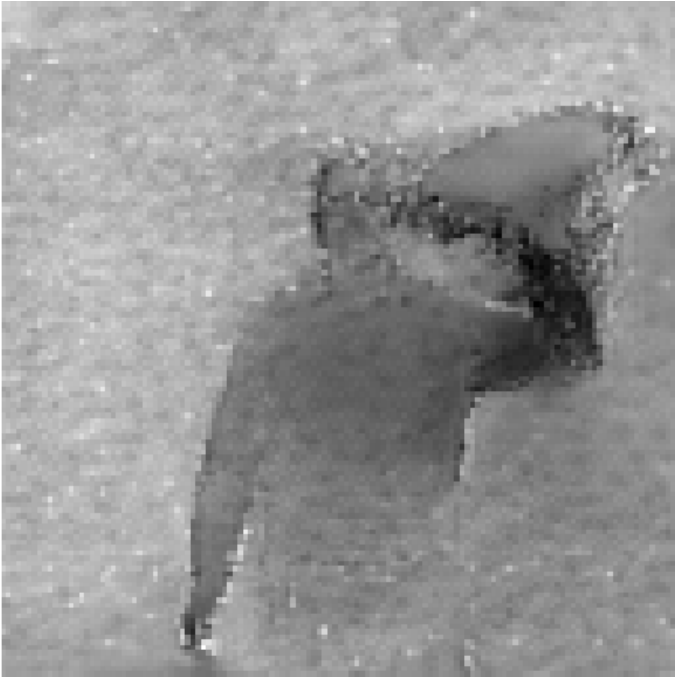}
  \end{subfigure}
  \begin{subfigure}[c]{0.185\textwidth}
    \includegraphics[width=\linewidth]{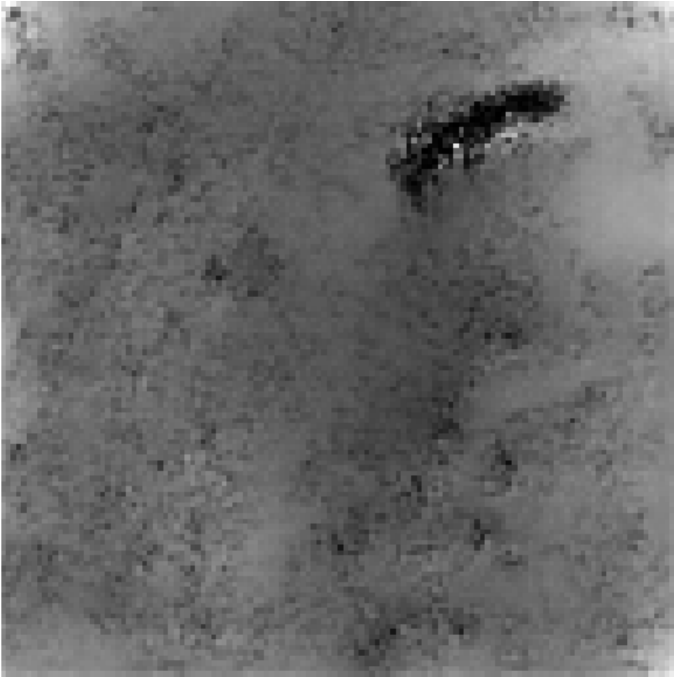}
  \end{subfigure}
  
  \vspace{0.1em}
  
  \begin{subfigure}[c]{0.03\textwidth}
    \rotatebox[origin=c]{90}{DVS-Gesture}
  \end{subfigure}%
  \begin{subfigure}[c]{0.185\textwidth}
    \includegraphics[width=\linewidth]{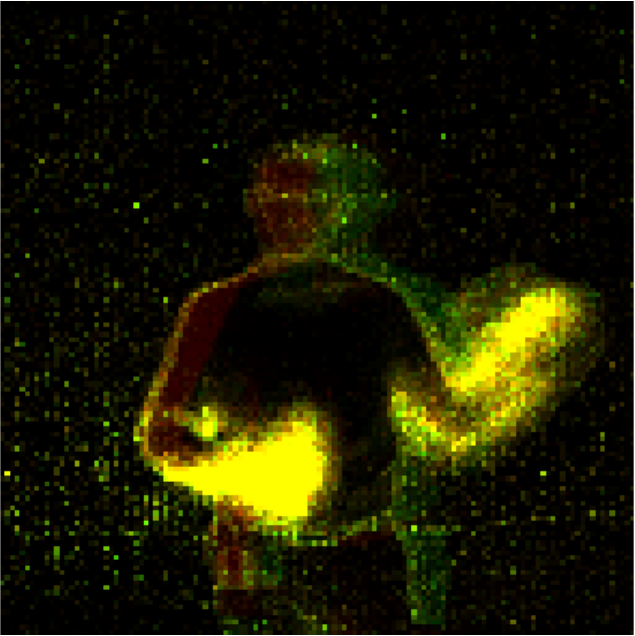}
  \end{subfigure}
  \begin{subfigure}[c]{0.185\textwidth}
    \includegraphics[width=\linewidth]{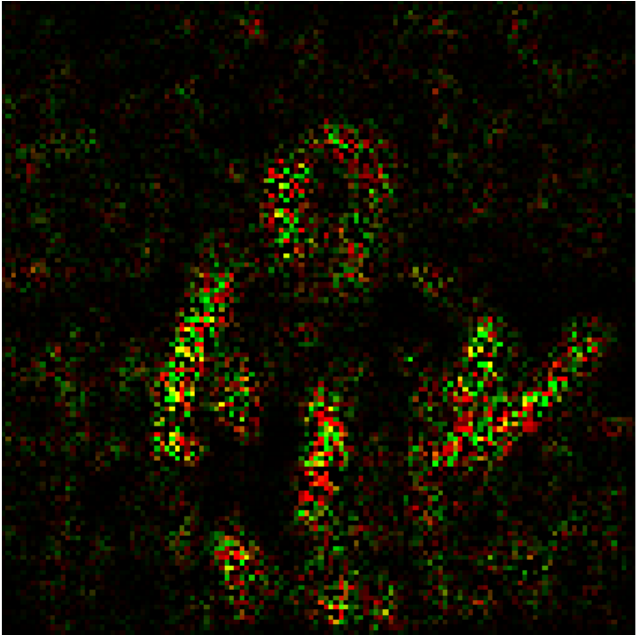}
  \end{subfigure}
  \begin{subfigure}[c]{0.185\textwidth}
    \includegraphics[width=\linewidth]{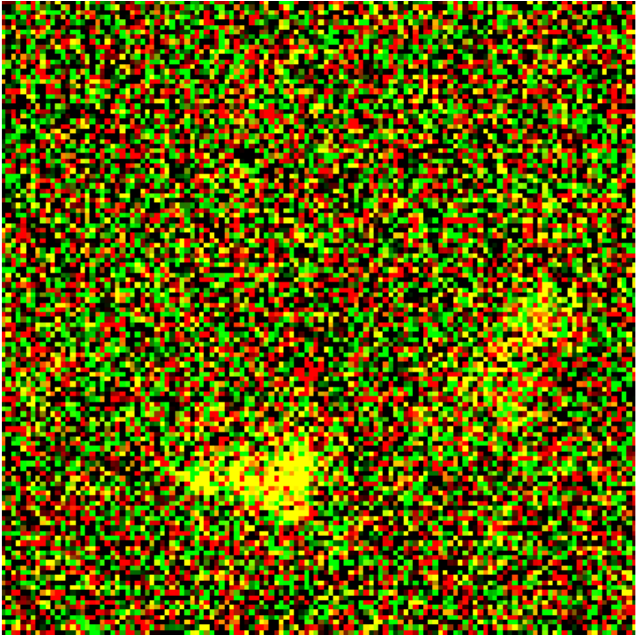}
  \end{subfigure}
  \begin{subfigure}[c]{0.185\textwidth}
    \includegraphics[width=\linewidth]{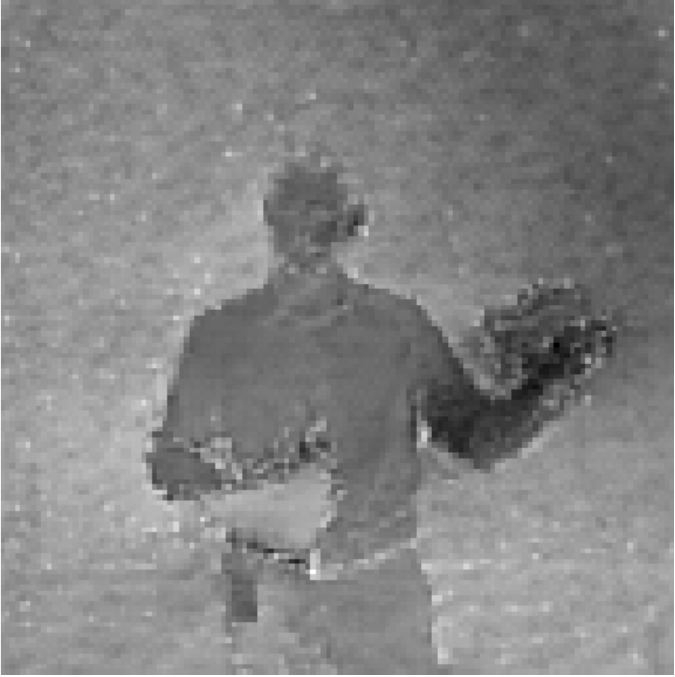}
  \end{subfigure}
  \begin{subfigure}[c]{0.185\textwidth}
    \includegraphics[width=\linewidth]{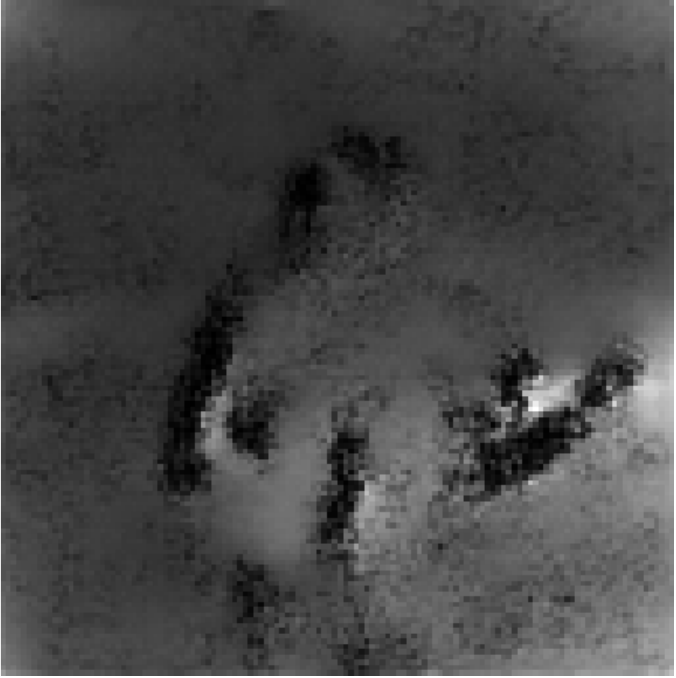}
  \end{subfigure}
  
  \vspace{0.1em}
  
  \begin{subfigure}[c]{0.03\textwidth}
    \rotatebox[origin=c]{90}{SEE}
  \end{subfigure}%
  \begin{subfigure}[c]{0.185\textwidth}
    \includegraphics[width=\linewidth]{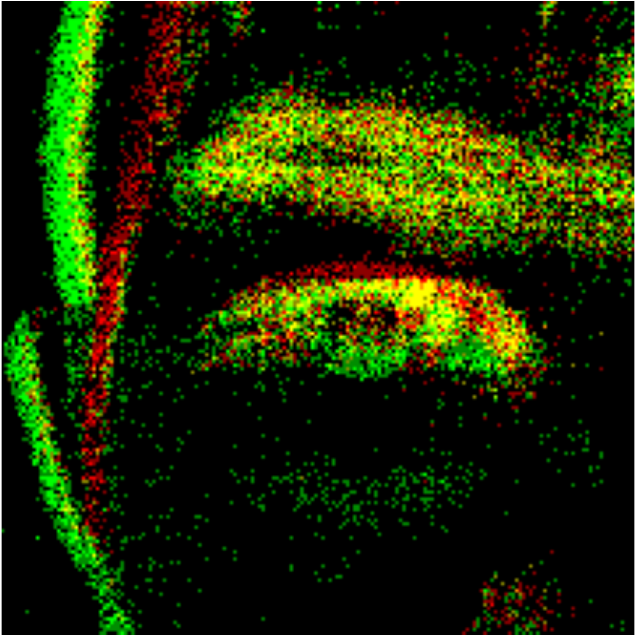}
  \end{subfigure} 
  \begin{subfigure}[c]{0.185\textwidth}
    \includegraphics[width=\linewidth]{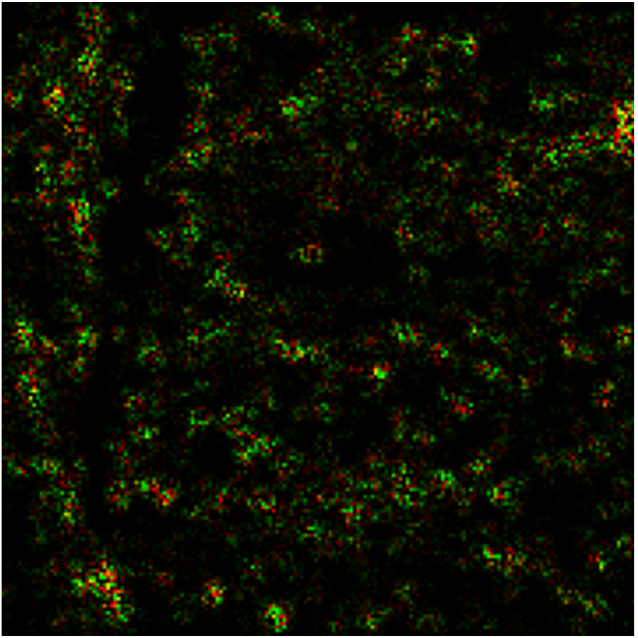}
  \end{subfigure}
  \begin{subfigure}[c]{0.185\textwidth}
    \includegraphics[width=\linewidth]{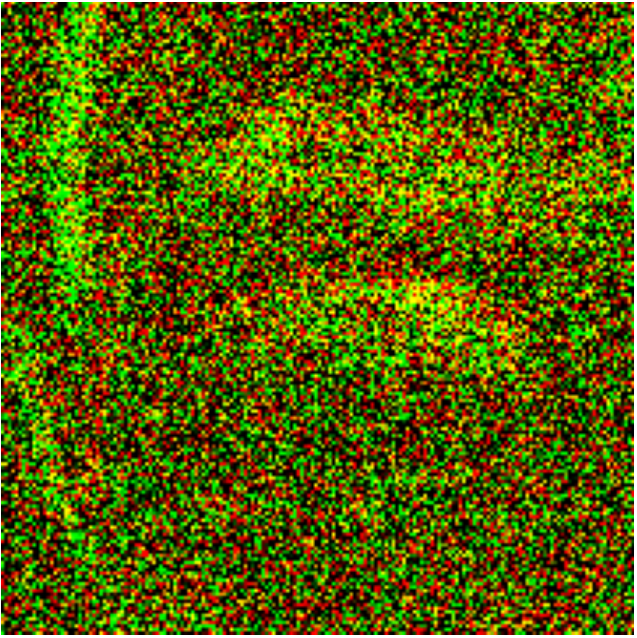}
  \end{subfigure}
  \begin{subfigure}[c]{0.185\textwidth}
    \includegraphics[width=\linewidth]{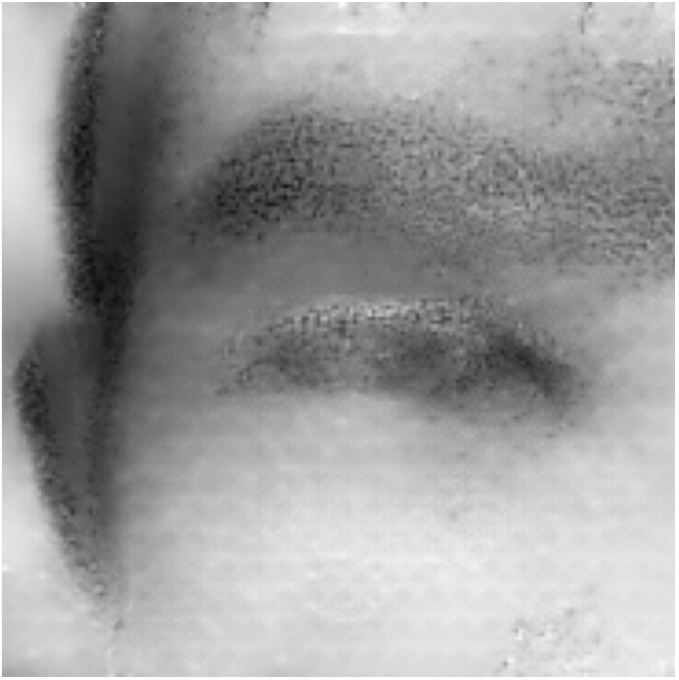}
  \end{subfigure}
  \begin{subfigure}[c]{0.185\textwidth}
    \includegraphics[width=\linewidth]{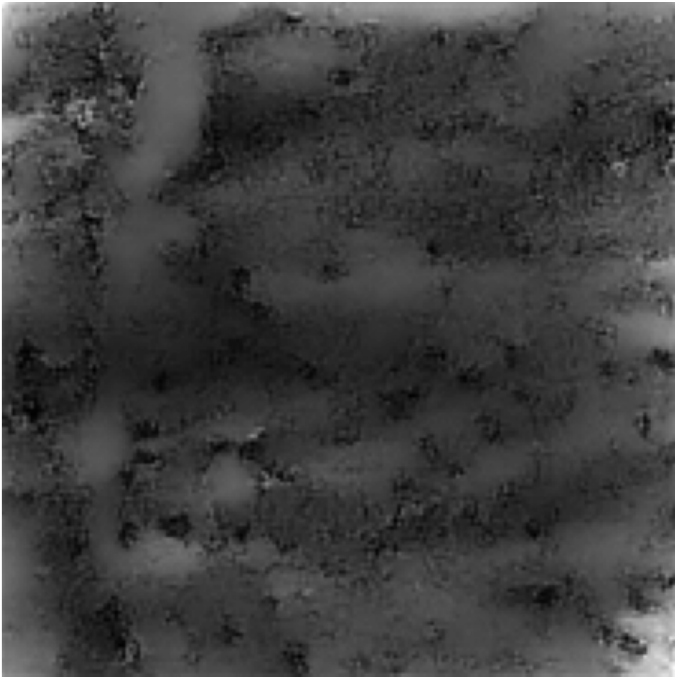}
  \end{subfigure}
  
 \vspace{0.1em}

  \begin{subfigure}[c]{0.03\textwidth}
    \rotatebox[origin=c]{90}{SEE}
  \end{subfigure}%
  \begin{subfigure}[c]{0.185\textwidth}
    \includegraphics[width=\linewidth]{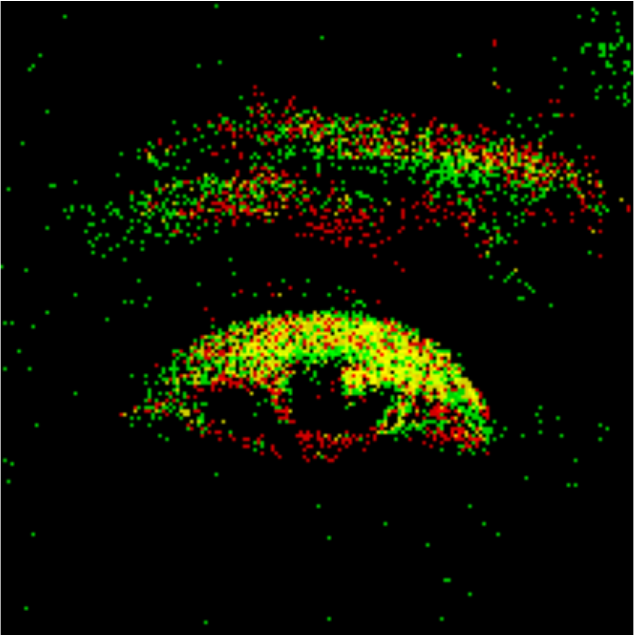}
  \end{subfigure}
  \begin{subfigure}[c]{0.185\textwidth}
    \includegraphics[width=\linewidth]{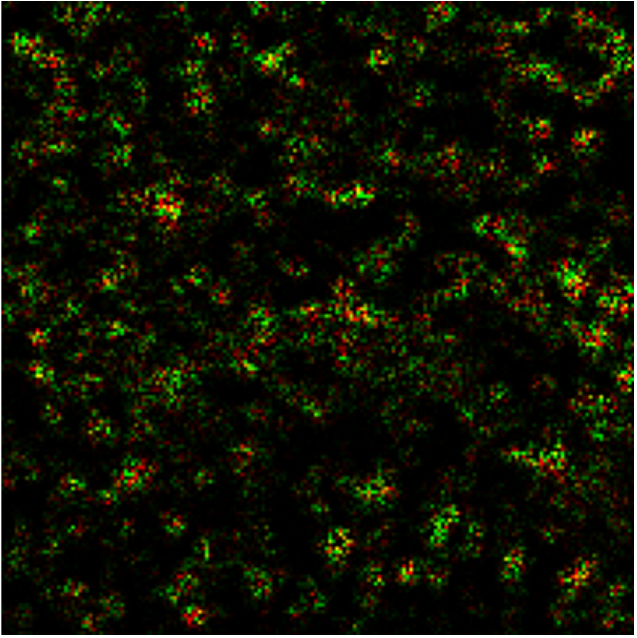}
  \end{subfigure}
  \begin{subfigure}[c]{0.185\textwidth}
    \includegraphics[width=\linewidth]{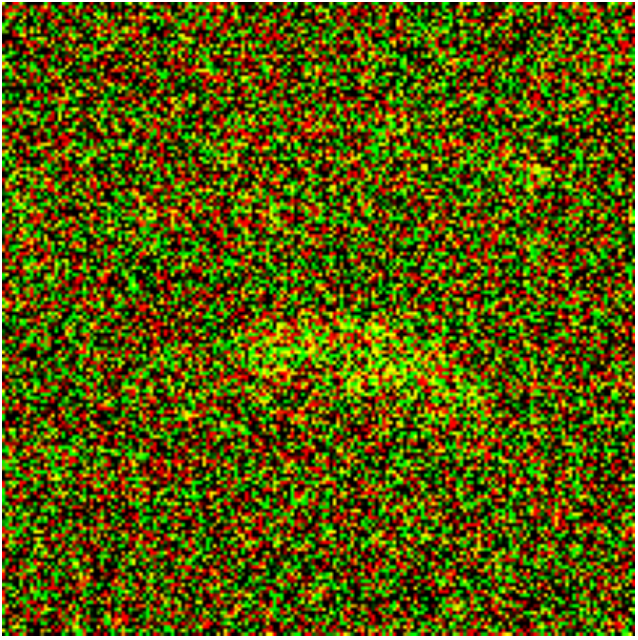}
  \end{subfigure}
  \begin{subfigure}[c]{0.185\textwidth}
    \includegraphics[width=\linewidth]{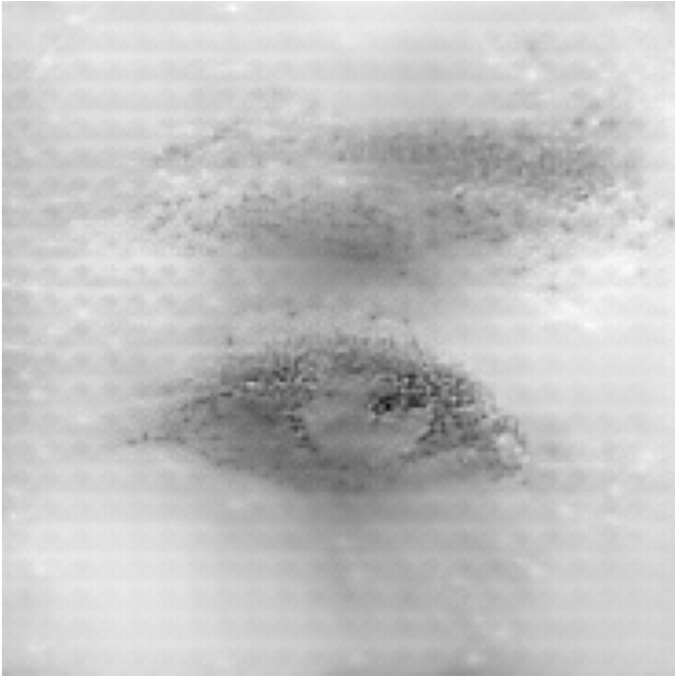}
  \end{subfigure}
  \begin{subfigure}[c]{0.185\textwidth}
    \includegraphics[width=\linewidth]{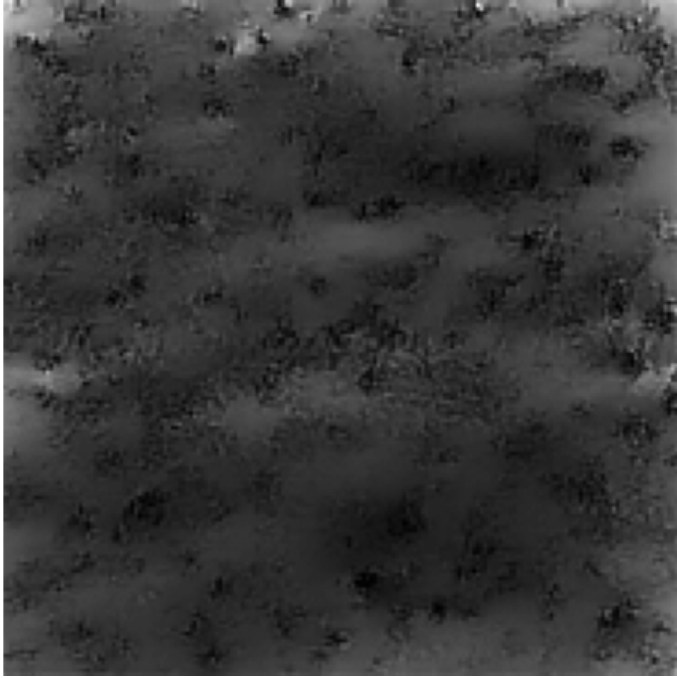}
  \end{subfigure}
  
  \vspace{0.1em}
  
  \begin{subfigure}[c]{0.03\textwidth}
    \rotatebox[origin=c]{90}{Re-ID}
  \end{subfigure}%
  \begin{subfigure}[c]{0.185\textwidth}
    \includegraphics[width=\linewidth]{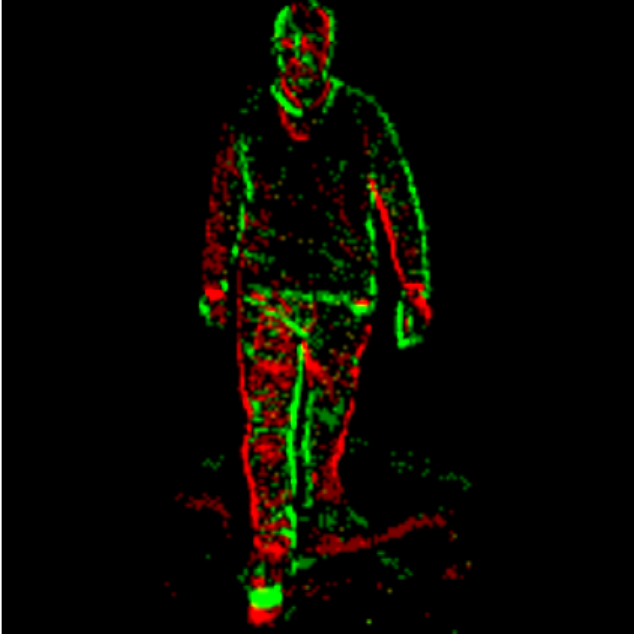}
  \end{subfigure}
  \begin{subfigure}[c]{0.185\textwidth}
    \includegraphics[width=\linewidth]{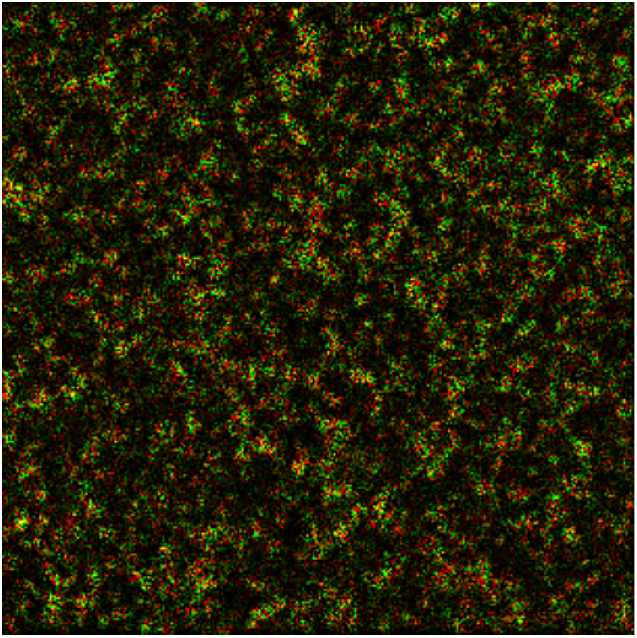}
  \end{subfigure}
  \begin{subfigure}[c]{0.185\textwidth}
    \includegraphics[width=\linewidth]{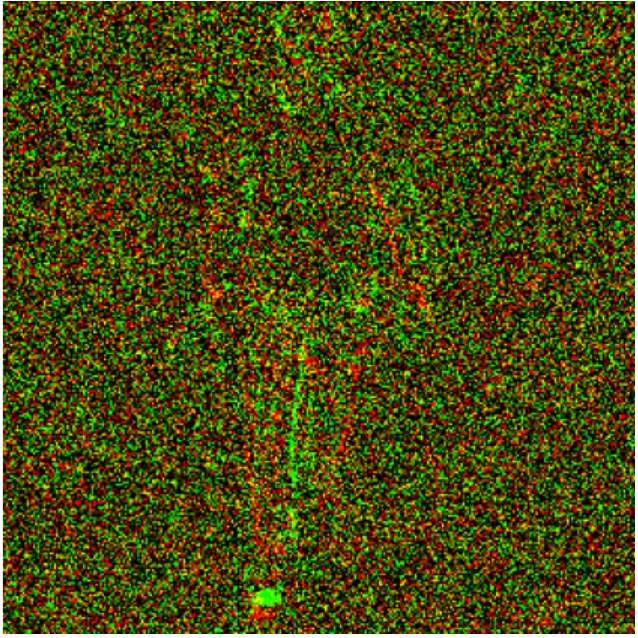}
  \end{subfigure}
  \begin{subfigure}[c]{0.185\textwidth}
    \includegraphics[width=\linewidth]{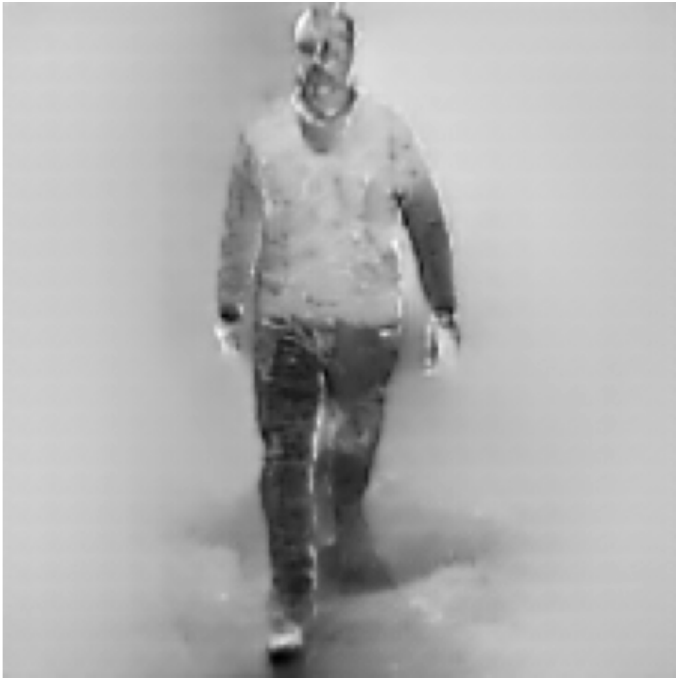}
  \end{subfigure}
  \begin{subfigure}[c]{0.185\textwidth}
    \includegraphics[width=\linewidth]{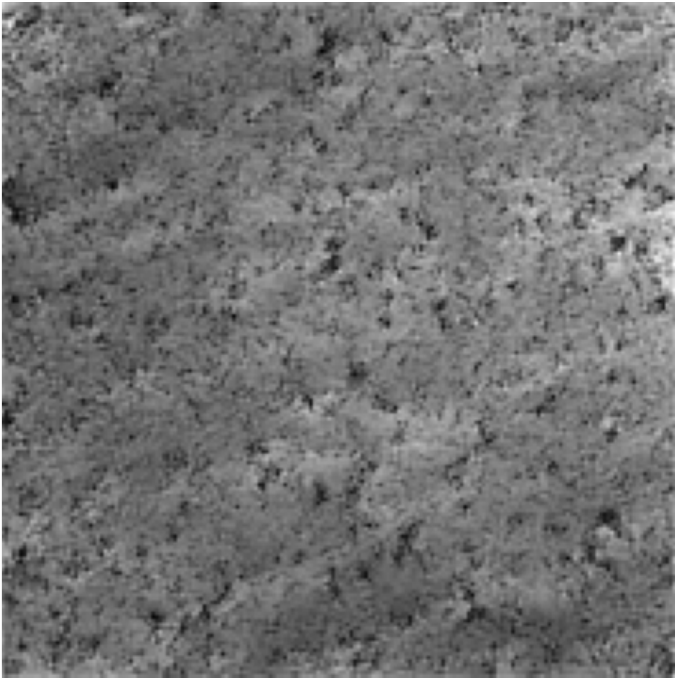}
  \end{subfigure}
  
  \vspace{0.1em}
  
  \begin{subfigure}[c]{0.03\textwidth}
    \rotatebox[origin=c]{90}{Re-ID}
  \end{subfigure}%
  \begin{subfigure}[c]{0.185\textwidth}
    \includegraphics[width=\linewidth]{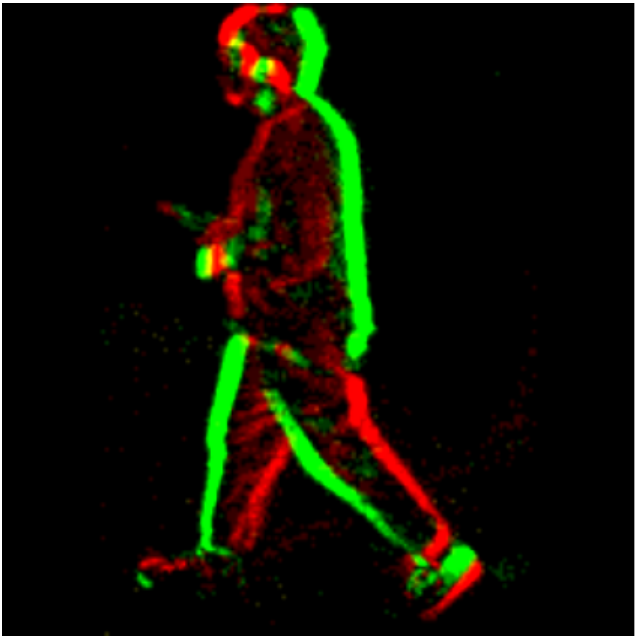}
    \caption{Original}
  \end{subfigure} 
  \begin{subfigure}[c]{0.185\textwidth}
    \includegraphics[width=\linewidth]{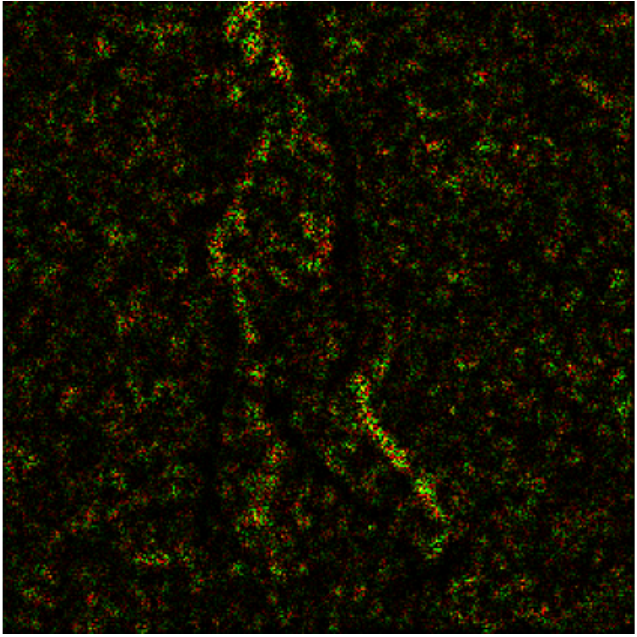}
    \caption{AnonyNoise}
  \end{subfigure}
  \begin{subfigure}[c]{0.185\textwidth}
    \includegraphics[width=\linewidth]{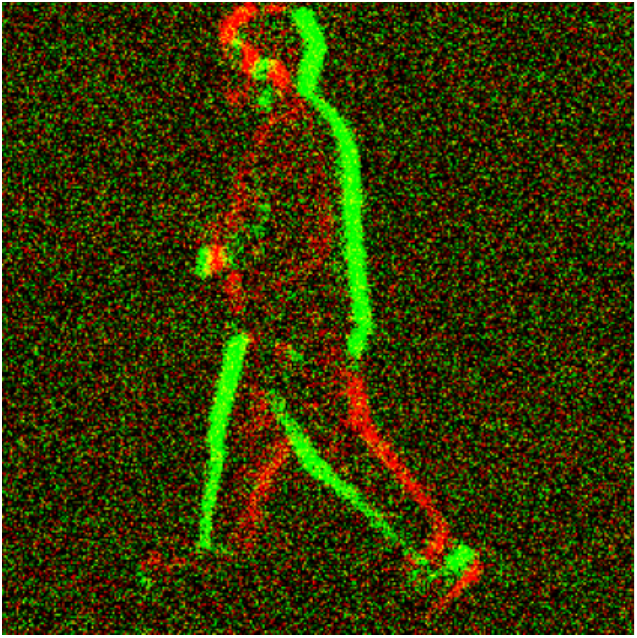}
    \caption{Gaussian Noise}
  \end{subfigure}
  \begin{subfigure}[c]{0.185\textwidth}
    \includegraphics[width=\linewidth]{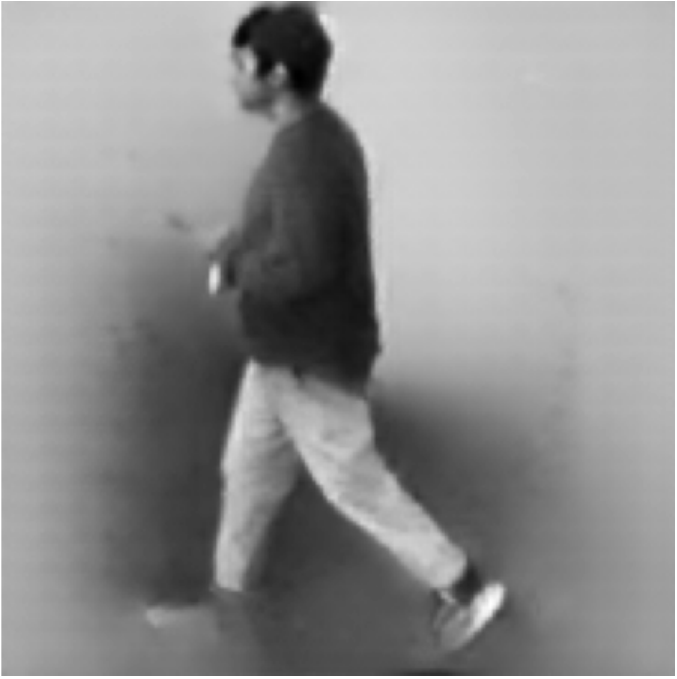}
    \caption{Original Recon.}
  \end{subfigure}
  \begin{subfigure}[c]{0.185\textwidth}
    \includegraphics[width=\linewidth]{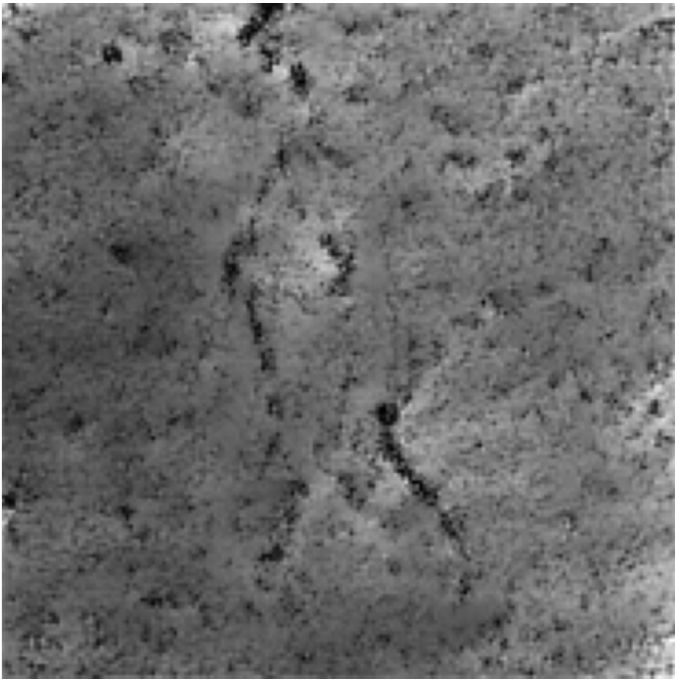}
    \caption{AnonyNoise Recon.}
  \end{subfigure}
  
  \vspace{0.1em}
  \caption{Visualized examples of a) the  original event data, b) the anonymized event, c) original events with Gaussian noise, d) grayscale image reconstruction based on the original event data and e) grayscale image reconstruction based on the anonymized events for DVS-Gesture \cite{dvsgesture}, SEE \cite{see}, and EventReId \cite{ahmad2022event}. The image reconstruction is based on E2VID \cite{rebecq2019high}. We apply Gaussian noise with a standard deviation of 32 for DVS-Gesture and of 1 for the other datasets. The images are visually enhanced for human perception.}
  \label{fig:vis_events_imagerecon_inversion}
\end{figure*}

In \cref{fig:vis_events_imagerecon_inversion}, we present a broader range of visual examples from the three datasets: DVS-Gesture \cite{dvsgesture}, SEE \cite{see}, and EventReId \cite{ahmad2022event}. The visualizations include raw and anonymized event data, as well as grayscale image reconstructions from both raw and anonymized events. Since EventReId contains only re-identification labels, we leverage the weights trained on DVS-Gesture for our anonymization network.

The work of \cite{ahmad2022event} specifically addresses the threat of grayscale image reconstruction from event data, which enables easy human identification. Their method employs an explicit loss during training to limit the reconstruction of images from anonymized events, while still allowing re-identification.

In contrast, our AnonyNoise method prevents re-identification by both humans and neural networks without any explicit loss preventing image reconstruction. To show this, we use the same pre-trained E2VID network \cite{rebecq2019high} employed in \cite{ahmad2022event}. E2VID, based on a recurrent neural network architecture, is capable of producing high-quality grayscale reconstructions from event data. However, the qualitative results of the image reconstruction attack, shown in \cref{fig:vis_events_imagerecon_inversion}, demonstrate that the attack failed for all three datasets. The images confirm that no human-recognizable reconstructions can be achieved, proving that our method is robust against such attacks. Furthermore, the results on EventReId show that AnonyNoise generalizes well even to previously unseen data.

\end{document}